\documentclass[10pt, a4paper]{article}

\usepackage{lrec2026} 

\usepackage[colorinlistoftodos,prependcaption,textsize=tiny]{todonotes}
\usepackage{dirtytalk}
\usepackage{tikz}
\usetikzlibrary{chains, positioning, fit, arrows.meta, shapes.geometric, calc}
\usepackage{dsfont}
\usepackage{multirow}
\usepackage{enumitem}
\usepackage{bbold}
\usepackage{tabularx}
\usepackage{etoolbox}
\usepackage[mathscr]{euscript}
\usepackage{tcolorbox}
\usepackage{amsmath}
\usepackage{xcolor}
\usepackage{textcomp} 
\usepackage{caption}
\usepackage{subcaption} 
\usepackage{booktabs}

\tcbuselibrary{listings, skins, breakable}
\newtcbox{\speakerbox}{on line, colback=orange!10, colframe=orange!60!black, boxrule=0.3pt, arc=1pt, boxsep=0.5pt, left=1pt, right=1pt, top=0.5pt, bottom=0.5pt}
\newtcbox{\actionbox}{on line, colback=blue!10, colframe=blue!60!black, boxrule=0.3pt, arc=1pt, boxsep=0.5pt, left=1pt, right=1pt, top=0.5pt, bottom=0.5pt}
\newtcbox{\cotimebox}{on line, colback=green!10, colframe=green!60!black, boxrule=0.3pt, arc=1pt, boxsep=0.5pt, left=1pt, right=1pt, top=0.5pt, bottom=0.5pt}
\newtcbox{\interlocutorbox}{on line, colback=magenta!10, colframe=magenta!60!black, boxrule=0.3pt, arc=1pt, boxsep=0.5pt, left=1pt, right=1pt, top=0.5pt, bottom=0.5pt}

\DeclareMathOperator{\sr}{\mathit{sr}}

\DeclareMathOperator{\ap}{\mathscr{A}_\mathbb{1}}
\DeclareMathOperator{\an}{\mathscr{A}_\mathbb{0}}
\DeclareMathOperator*{\argmax}{arg\,max}

\newcommand\blfootnote[1]{%
  \begingroup
  \renewcommand\thefootnote{}\footnote{#1}%
  \addtocounter{footnote}{-1}%
  \endgroup
}

\newcommand{\fn}{FrameNet}

\newcommand{\sota}{state-of-the-art}

\title{FrameNet Semantic Role Classification by Analogy}

\name{Van-Duy Ngo$^{1,\dag}$, Stergos Afantenos$^{1,\dag}$, Emiliano Lorini$^{1}$, Miguel Couceiro$^2$} 

\address{$^1$IRIT, CNRS, INP, Université de Toulouse \\
$^2$INESC-ID, IST, Universidade de Lisboa\\
\{van-duy.ngo, emiliano.lorini, stergos.afantenos\}@irit.fr\\
miguel.j.couceiro@tecnico.ulisboa.pt\\}

\abstract{%
In this paper, we adopt a relational view of analogies applied to Semantic Role Classification in \fn{}.
We define analogies as formal relations over the Cartesian product of frame evoking lexical units (LUs) and frame element (FEs) pairs, which we use to construct a new dataset.
Each element of this binary relation is labelled as a valid analogical instance if the frame elements share the same semantic role, or as invalid otherwise.
This formulation allows us to transform Semantic Role Classification into binary classification and train a lightweight Artificial Neural Network (ANN) that exhibits rapid convergence with minimal parameters.
Unconventionally, no Semantic Role information is introduced to the neural network during training. 
We recover semantic roles during inference by computing probability distributions over candidates of all semantic roles within a given frame through random sampling and analogical transfer.
This approach allows us to surpass previous \sota{} results while maintaining computational efficiency and frugality.
 \\ \newline  }

\begin{document}
\maketitleabstract
\section{Introduction}\label{sec:intro}
Analogical\blfootnote{Authors with $\dag$ are the main contributors.} reasoning is a human faculty that is ``at the core of cognition'' \cite{Hofstadter.01} and although this might not be an opinion that is widely shared in the research community at large, it is undeniable that recent advances regarding Large Language Models (LLMs) have sparked a new interest in the research community for analogies, including the field of Natural Language Processing (NLP). 
More recently, \citet{Firt:analogical:2025} argues that reaching what has come to be known as Artificial General Intelligence (AGI) will only become possible if analogies serve as a fundamentally basic component of such systems. 
Despite the diachronic interest in analogies, it seems to be the case that by the term ``analogies'', different disciplines refer to varying viewpoints or even interpretations, since this term is used in fields ranging from cognitive science \cite{Gentner.83,hofstadter_copycat_1984,holyoak_mental_1995} and computational linguistics \cite{MikolovNIPS2013,drozd-etal-2016-word} to formal logic \cite{prade2009analogy} and machine learning \cite{Bounhas.Prade.2023,CunhaAAAI2026}.

Most approaches in NLP adopt the proportional view of analogies~\cite{Aristotle2009} of form $a:b::c:d$, which reads \say{$a$ is to $b$ as $c$ is to $d$}, with $a, b, c, d$ being abstract entities (\textit{e.g.} boolean variables, vectors, etc) and $a:b::c:d$ representing a valid analogy if the same transformations (usually taking the form of a difference) that happen from $a$ to $b$ also hold from $c$ to $d$
\cite{barbot_analogy_2019,DBLP:conf/iccbr/PradeR17}. 
From a cognitive science perspective, \citet{Gentner.83} establishes the Structured Mapping Theory (SMT), which emphasises the rather fundamental role of structural relationships among components than their attributes in forming relational analogies.
\citet{hofstadter_fluid_1995,mitchell_analogy_1993,chalmers_high-level_1992,hofstadter_surfaces_2013} argue that analogies is a ubiquitous and fluid mechanism involved in the formation of concepts. 
As such, it can be highly dependent on the context, in the sense that an object $a$ can be considered analogous to object $b$ only under a specific context. 

In NLP, most researchers adopt the proportional point of view of analogies \cite{efficient-representation-w2v:2013:mikolov,MikolovNIPS2013,drozd-etal-2016-word}
and mainly focus on analogies at lexical levels, such as the lexicons in the well-cited $\mathtt{king:man::queen:woman}$ example of \citet{MikolovNIPS2013}. 
Typically, words in the quadruplet are represented by either static or contextual vector embeddings, and the quadruplet's validity is realised by arithmetic or geometric differences.
The relation that holds between pairs $(a,b)$ and $(c,d)$ is usually categorized into either inflectional, derivational, lexicographic, or encyclopedic relations \cite{drozd-etal-2016-word}. 
Proportional views of analogies have also been applied for morphosyntactic patterns such as $\mathtt{run:running::drive:driving}$ \cite{lepage_translation_2005,ulcar-etal-2020-multilingual,karpinska-etal-2018-subcharacter,MarquerC25}. 
For languages that do not exhibit complex morphosyntactic patterns,\footnote{Some non-agglutinative languages, for example.} such a task is rather trivial \cite{ushio-etal-2021-bert,petersen-van-der-plas-2023-language}. 

Despite its usefulness for certain tasks, proportional approaches of analogies in Natural Language assume that the same common latent relation always holds between the two pairs of words that constitute a valid analogy, independently of anything else. 
This misses a crucial aspect regarding the pragmatics of Natural Language namely, the fact that such latent relations can dramatically shift depending on the \emph{context} in which the lexical items of the quadruplet are found in.\footnote{This remark concerns primarily the construction of datasets, prior to any computational attempt in recognising analogies. It has to be noted that most extant approaches use one form or another of distributional semantics \cite{firth1968selected,harris1954distributional} and thus implicit analogy recognition relies on the context.} 
Take for example the observation from \citeauthor{mickus-etal-2023-mann} on the fact that the pairs $\mathtt{flower:petal}$ and $\mathtt{tonne:kilogram}$ both appear in the L06 meronym category of the BATS corpus \cite{drozd-etal-2016-word} thus forming an analogy, despite 
that the in-context validity of such analogy is very much debatable.  

In this paper, we do not strictly subscribe to the proportional approach to analogies, but instead get inspired by the aforementioned theories in Cognitive Science. 
We propose that analogical validity in NLP---whether a quadruplet of textual elements\footnote{Expressed as a couple of pairs, in our case.} forms a valid analogy---depends vastly on the latent semantic relations established between those elements, which are themselves heavily influenced by their contextual environments. 
This highlights the fundamental principle that semantic interpretation is irrefutably shaped by pragmatic factors.

To demonstrate our point, consider the following quadruplet: 
$\mathtt{flower:petal::tree:leaf}$.
On a lexical level, the meronymous relations are apparent hence the positive validity.
Let us now place those four words in a context:
\begin{enumerate}[label*=(\arabic*),series=example]\footnotesize
  \item{Decomposed leaves provide a fertilizer for the trees.\label{ex:decleaves}}
  \item{Infected petals imperil the life-span of the flowers.\label{ex:infpetals}}
\end{enumerate}
In this context, the holonym-meronym relations between $\mathtt{flower:petal}$ and $\mathtt{tree:leaf}$ become much less relevant. 
Instead, the semantics of the two sentences play a more prominent role in judging whether an analogy holds for this quadruplet. 
In this case, we are inclined to conclude that the lexical analogy no longer holds, since in~\ref{ex:decleaves} the decomposed leaves are beneficial for the trees while in~\ref{ex:infpetals} infected petals are harmful for the flowers, completely eclipsing the holonym-meronym relations. 
This is well reflected on the frame semantics of the two sentences, using \fn{} \cite{baker-etal-1998-berkeley-framenet,bakerarticle2017framenet} annotations:
\begin{enumerate}[label*=(\arabic*),resume*=example]
\item{\textit{Frame: Supply} \ \\
  \scalebox{0.7}{%
  \begin{tikzpicture}[
    node distance = 1mm,
    start chain = A going right,
    txt/.style = {text height=2ex, text depth=0.01ex, on chain},
    every edge/.append style = {draw, -syellowth'},
    every node/.append style = {very thick,rounded corners=10pt}
    ]
    \node [txt] {Decomposed leaves};     
    \node [txt] {provide};
    \node [txt] {a fertilizer};
    \node [txt] {for};
    \node [txt] {the trees};
    \node (f1)  [draw=red, fill=red!20, fill opacity=0.2,inner sep=0pt, 
                 label=below:\textsc{Supplier}, 
                 fit=(A-1) ] {};
    \node (f2)  [draw=black,inner sep=0pt, 
                 label=below:predicate,
                  fit=(A-2) ] {};
    \node (f2)  [draw=yellow,fill=yellow!20, fill opacity=0.2,inner sep=0pt, 
                 label=below:\textsc{Theme},
                  fit=(A-3) ] {};
    \node (f2)  {};
    \node (f2)  [draw=green,fill=green!20, fill opacity=0.2,inner sep=0pt, 
                 label=below:\textsc{Recipient},
                  fit=(A-5) ] {};
\end{tikzpicture}}
\label{ex:decleaves_FN}}
\item{
\textit{Frame: Endangering} \ \\
  \scalebox{0.7}{%
  \begin{tikzpicture}[
node distance = 1mm,
  start chain = A going right,
   txt/.style = {text height=2ex, text depth=0.01ex,
                 on chain},
every edge/.append style = {draw, -syellowth'},
every node/.append style = {very thick,rounded corners=10pt}
                        ]
\node [txt] {Infected petals};     
\node [txt] {imperil};
\node [txt] {the life-span of the flowers};
\node (f1)  [draw=blue, fill=blue!20, fill opacity=0.2,inner sep=0pt, 
             label=below:\textsc{Cause}, 
             fit=(A-1) ] {};
\node (f2)  [draw=black,inner sep=0pt, 
             label=below:predicate,
              fit=(A-2) ] {};
\node (f2)  [draw=magenta,fill=magenta!20, fill opacity=0.2,inner sep=0pt, 
             label=below:\textsc{Valued Entity},
              fit=(A-3) ] {};
    \end{tikzpicture}
    }
\label{ex:infpetals_FN}}
\end{enumerate}
Suppose now that we have the following sentence annotated with \fn{} annotations:
\begin{enumerate}[label*=(\arabic*),resume*=example]
\item{
\textit{Frame: Supply} \ \\ 
  \scalebox{0.68}{%
  \begin{tikzpicture}[
    node distance = 1mm,
    start chain = A going right,
    txt/.style = {text height=2ex, text depth=0.01ex,
                 on chain},
    every edge/.append style = {draw, -syellowth'}, every node/.append style = {very thick,rounded corners=10pt}
                            ]
    \node [txt] {Blossoming petals};     
    \node [txt] {supply};
    \node [txt] {precious pollinators};
    \node [txt] {to};
    \node [txt] {flowers};
    \node (f1)  [draw=red, fill=red!20, fill opacity=0.2,inner sep=0pt, 
                 label=below:\textsc{Supplier}, 
                 fit=(A-1) ] {};
    \node (f2)  [draw=black,inner sep=0pt, 
                 label=below:predicate,
                  fit=(A-2) ] {};
    \node (f2)  [draw=yellow,fill=yellow!20, fill opacity=0.2,inner sep=0pt, 
                 label=below:\textsc{Theme},
                  fit=(A-3) ] {};
    \node (f2)  {};
    \node (f2)  [draw=green,fill=green!20, fill opacity=0.2,inner sep=0pt, 
                 label=below:\textsc{Recipient},
                  fit=(A-5) ] {};
    \end{tikzpicture}}
\label{ex:blosspetals_FN}}
\end{enumerate}
Although no analogy can be formed between the elements of sentences~\ref{ex:decleaves_FN} and~\ref{ex:infpetals_FN}, we can discern an analogy on the semantic level between~\ref{ex:decleaves_FN} and~\ref{ex:blosspetals_FN} in the sense that both $\mathtt{leaves}$ and $\mathtt{petals}$ are meant to \emph{supply} something to $\mathtt{trees}$ and $\mathtt{flowers}$ respectively. 
This is reinforced by the fact that the focused predicates trigger the same frame \emph{Supply} and the frame-role relations within the two predicate-frame element pairs remain identical.

Let us now consider the following quadruplet:
$\mathtt{leaves:trees::lentils:humans}$.
No obvious common relation, of meronymy or otherwise, can be discerned between the two pairs. 
Without further context, we cannot meaningfully claim that this quadruplet forms a valid analogy since we cannot say that \say{\textit{leaves} are to \textit{trees} as \textit{lentils} are to \textit{humans}}. 
However, when we consider the following sentence:
\begin{enumerate}[label*=(\arabic*),resume*=example]
\item{
\textit{Frame: Supply} \ \\
  \scalebox{0.7}{%
  \begin{tikzpicture}[
    node distance = 1mm,
    start chain = A going right,
    txt/.style = {text height=2ex, text depth=0.01ex, on chain},
    every edge/.append style = {draw, -syellowth'}, every node/.append style = {very thick,rounded corners=10pt}
                        ]
    \node [txt] {Lentils\ \ \ \ \ };     
    \node [txt] {furnish};
    \node [txt] {proteins};
    \node [txt] {to};
    \node [txt] {humans};
    \node (f1)  [draw=red, fill=red!20, fill opacity=0.2,inner sep=0pt, 
                 label=below:\textsc{Supplier}, 
                 fit=(A-1) ] {};
    \node (f2)  [draw=black,inner sep=0pt, 
                 label=below:predicate,
                  fit=(A-2) ] {};
    \node (f2)  [draw=yellow,fill=yellow!20, fill opacity=0.2,inner sep=0pt, 
                 label=below:\textsc{Theme},
                  fit=(A-3) ] {};
    \node (f2)  {};
    \node (f2)  [draw=green,fill=green!20, fill opacity=0.2,inner sep=0pt, 
                 label=below:\textsc{Recipient},
                  fit=(A-5) ] {};
\end{tikzpicture}}
\label{ex:lentils_FN}}
\end{enumerate}
there is obviously a semantic analogy between~\ref{ex:decleaves_FN} and~\ref{ex:lentils_FN}. 

Inspired by cognitive theories of analogy, particularly the approach put forth by 
\citet{hofstadter_surfaces_2013} as well as the Structure Mapping Theory~\cite{Gentner.83,gentner_hoyos:2017}, we formulate analogies based on the semantic roles that textual elements hold in relation to predicate words that trigger their containing frames.
For instance, in sentences~\ref{ex:decleaves_FN} and~\ref{ex:lentils_FN}, we identify the following quadruplets $a:b::c:d$ as valid analogies:

\medskip\noindent
\resizebox{\linewidth}{!}{$\mathtt{provide:decomposed\ leaves::furnish:lentils}$\\}
\resizebox{0.91\linewidth}{!}{$\mathtt{provide:a\ fertilizer::furnish:proteins}$\\}
\resizebox{0.79\linewidth}{!}{$\mathtt{provide:the\ trees::furnish:humans}$}

\medskip\noindent
As \citet{petersen_et_al:2025} observe in their review of analogical reasoning at the intersection of NLP and cognitive science, this directly aligns with the \textit{mapping} process in SMT: elements $b$ and $d$ playing the same semantic role within the common semantic frame triggered by \emph{anchor} elements $a$ and $c$ constitute a valid analogy, while all other combinations with semantic role mismatch do not. 

Constructing
pairs of
predicate-arguments
as described above allows us to build a novel dataset of positive and negative instances based on \fn{} annotations, from which we can train a model to learn the validity of such analogies. 
We subsequently use the model to identify pairs whose frame element's semantic role is analogous to the input, which corresponds to the \textit{retrieval} process in SMT.
In our case, the method is as follows: given a sentence with spans representing frame elements to be classified, we sample random annotated sentences corresponding to the same semantic frame and build quadruplets that we submit to a trained analogical model, assigning to each span probability distributions over all samples of semantic roles of the frame, then grant the frame element the semantic role with the highest probability.

The structure of the paper is as follows: in \S\ref{sec:fn} we briefly introduce \fn{} and Semantic Role Classification, while in \S\ref{sec:fn_an} we present in more detail how we construct analogical quadruplets from \fn{} annotations, which corresponds to the \textit{mapping} phase of SMT, in order to create a training set with positive and negative instances for training neural network models. 
In the same section, we describe how this model can be used during inference 
for Semantic Role Classification. 
\S\ref{sec:expe} presents the neural network architecture 
, the experimental settings, and the results obtained. 
Related work is briefed in \S\ref{sec:rel_work}.
After discussing various aspects of this work and the prospect of applying analogy to diverse tasks in \S\ref{sec:disc}, we conclude.

\section{FrameNet and Semantic Role Classification}\label{sec:fn}
The Berkeley \fn{} project \cite{baker-etal-1998-berkeley-framenet,bakerarticle2017framenet,ruppenhofer2006extended} is based on the theo\-ry of Frame Semantics established by \citet{fillmore1976frame}. 
The main idea behind \fn{} is that the realisation of word's meaning requires a semantic {frame} that is {evoked} by a trigger.
A semantic frame consists of a predicate---usually a verb, noun, or adjective---that triggers the frame, along with several semantic roles that serve as arguments to the frame. 
Semantic roles are either 
\emph{core} that are almost always present in a sentence\footnote{Niche grammatical constructions do not allow instantiation of some core elements. In such cases a NI (Not Instantiate) label is provided. Non-instantiations are divided into Definite (DNI) Indefinite (DNI) and Constructional (CNI).},
\emph{peripheral} whose instantiation is not mandatory, 
or \emph{extra-thematic} that capture information related to time, location, manner, means, degree, etc. 
While the latter are shared among many frames, core and peripheral roles are frame-specific.  
Semantic roles share a common ontological structure across all frame types.  
Figure~\ref{fig:fn_annotation} shows an example of an annotated sentence with frame triggered by the predicate \texttt{decline}.

\begin{figure}[htb]
    \noindent
    [\speakerbox{\scriptsize\textbf{Speaker}} Fenn] \textit{declined}\textsubscript{Trigger} 
    [\actionbox{\scriptsize\textbf{Proposed\_action}} the offer to buy] 
    [\cotimebox{\scriptsize\textbf{Co-timed\_event}} with a bemused wave of his hand]. 
    [\interlocutorbox{\scriptsize\textbf{Interlocutor}} DNI]  
        \caption{An example of an annotated sentence for the frame type triggered by \texttt{decline.v}.}
        \label{fig:fn_annotation}
\end{figure}

Frame Semantic Parsing \cite{gildea2002automatic,baker-etal-2007-semeval} is the process that aims to extract full frame semantic structures starting from a sentence.
The full process can be divided into three sub-tasks \cite{zheng-etal:2022}. 
\textbf{Frame Identification}: given a sentence, identify all the frames and their respective evoking predicates. 
\textbf{Argument Identification:} given a sentence, a frame, and the evoking predicate, identify all the elements in the sentence that represent arguments for that predicate. 
\textbf{Semantic Role Classification:} This sub-task takes as input the results of the previous two to assign the elements with corresponding semantic roles. 
On \fn{}, the task is termed \fn{} Parsing~\cite{palmer2010evaluating}.
Evaluation is empirically performed either for each separate module or collectively on the entire \fn{} Parsing pipeline. 
The prevalent former usually presents the results of a single sub-task on gold input rather than the output of the preceding module, posing compatibility problems for integration.
Integral pipelines for \fn{} Parsing---with modular components for each sub-task---emerged as the optimal solution for these problems~\cite{graph:2021:lin-etal,chanin2023opensource,Swayamdipta_etal_2017}.

\section{Learning Analogies for Semantic Role Classification}\label{sec:fn_an}
Our methodology to solving \fn{} Semantic Role Classification is inspired by cognitive science approaches to analogies, such as per \citet{hofstadter_surfaces_2013}, in which the context dictates the validity of an analogy. 
Our analysis roots for the use of predicates to encompass the context of the sentences, relying on (i) the theoretically sound activation of the frame by the predicate, 
and (ii) the frame-specific relations that come alongside the evoked frame.
This is advantageous because the same encoder can generate latent representations for both the context and the target frame element, thereby ensuring consistency and preventing representation mismatches.

By this principle, we build a dataset of analogical instances from \fn{} annotations and train a model for binary classification.
During training, the model receives no semantic role label information.
For inference, we construct a reference dataset of predicate-argument pairs by randomly selecting annotated instances that meet the frame constraint. 
Specifically, for every \textit{target} frame element $\mathtt{e^t}$ to classify with its known predicate $\mathtt{p^t}$, we randomly sample $n_\pi$ pairs of \textit{source} frame elements and associated predicates $\mathtt{p^{s}_i}:\mathtt{e^{s}_i}$ of the same frame type to build $n_\pi$ analogical instances of $\mathtt{p^{s}_i}:\mathtt{e^{s}_i}::\mathtt{p^t}:\mathtt{e^t}, 0<i\leq n_\pi$, then submit them to the trained model for classification.
Instances classified as positive are kept and the semantic role type of the source frame element is transferred to the target's.

To summarise, we divide our approach into three phases: 
\textbf{a)} creating a dataset of positive and negative instances representing analogies, \textbf{b)} training a neural network model for binary classification, and 
\textbf{c)} decoding of this model's output for multi-class classification via analogical transfer, based on obtained probability distributions over all semantic role types within a given frame. 

\subsection{Dataset Creation and Learning}
In order to train our binary analogical model, it is necessary to construct a dataset that contains positive and negative instances of analogies.  
Given a set $S$ of sentences annotated with Frame types and Semantic Roles according to \fn, we consider each sentence $s \in S$ as a sequence of tokens $s=\{\mathtt{t}_i\}_{i=1}^{l}$ with $l$ being the length of the sentence. 
For each sentence, along with the frame and the evoking predicate $\mathtt{p}$, we extract the frame elements $E = \{\mathtt{e_j}\}_{j=1}^{k}$, with $k$ representing the total number of frame elements. Each frame element $\mathtt{e_j}$ should be classified into its semantic role.
For each frame $\phi$ in the set of all frames $\Phi$,
we construct the set of all predicate-argument pairs $\mathtt{P}_\phi$. 
Note that a sentence $s$ may contain multiple triggers that by definition evoke multiple frames. 
Semantic roles for each frame element are obtained using function $\sr : \mathcal{E}\times\Phi \mapsto \mathcal{SR}_{\phi}$ with $\mathcal{E}$ representing the set of all frame elements and $\mathcal{SR}_{\phi}$ the set of all semantic role types for a given semantic frame $\phi \in \Phi$.

We then define two formal relations:\footnote{Verb diathesis and multiple valences of a verb could lead us to think that we should ``encode'' each verb as an n-ary relation. This is not what we try to do here. Instead we focus on a single argument for which we want to identify its semantic role.} $$\ap, \an \subseteq \bigcup_{\phi\in\Phi} \mathtt{P}_\phi \times \mathtt{P}_\phi$$ that respectively denote the positive/negative validity of an instance---which is a pair of predicate-arg\-ument pairs.
This allows for direct construction of a dataset containing positive and negative analogy instances. 
More precisely, given two pairs $\mathtt{(p_1,e_1)} \in \mathtt{P}_\phi, \mathtt{(p_2, e_2) \in \mathtt{P}_\phi}$ for any given $\phi \in \Phi$, the relation containing valid instances of analogies $\ap$, is defined as follows:
\begin{equation*}\footnotesize
     \mathtt{((p_1,e_1),(p_2,e_2))}\in\ap \Leftrightarrow \sr(\phi, \mathtt{e_1}) = \sr(\phi,\mathtt{e}_2)
     \label{eq:rel_ap}
\end{equation*}
Similarly, the set of quadruplets representing non-valid analogies is defined as:
\begin{equation*}\footnotesize
     \mathtt{((p_1,e_1),(p_2,e_2))}\in\an \Leftrightarrow \sr(\phi,\mathtt{e_1}) \neq \sr(\phi,\mathtt{e}_2)
     \label{eq:rel_an}
\end{equation*}
$\ap$ is an equivalence relation with equivalence classes representing semantic role types. $\an$ retains symmetry but not reflexivity or transitivity.

The definition of these two relations allows us to construct a set $\{(\mathbf{x}_i, y_i)\}_{i=1}^n$ of positive and negative instances.
More precisely: 
\begin{flalign*}
    & \forall\ \mathtt{((p_1,e_1),(p_2,e_2))} \in \ap \cup \an : & \nonumber \\ 
    \qquad & \mathbf{x} = [\mathtt{emb(p_1) \oplus emb(e_1) \oplus emb(p_2) \oplus emb(e_2) }], & \nonumber \\
    \qquad & y = \begin{cases}
        1, & \text{ if $\sr(\phi,\mathtt{e_1}) = \sr(\phi,\mathtt{e_2})$ } \\
        0, & \text{ otherwise}
    \end{cases} 
\label{eq:emb}
\end{flalign*}
where $\mathtt{emb(\lambda)}$ denotes an embedding function that provides a fixed-size vector for any text $\mathtt{\lambda}$, and $\oplus$ denotes concatenation. Quadruplets that belong to $\ap$ provide positive instances while those falling under $\an$ provide negative ones.
These instances are intended for the binary classifier regardless of their analogical validity.
Crucially, we never \emph{explicitly} provide semantic role information to the model during training.
This binary classifier $\mathcal{A} : \ap \cup \an \mapsto \{0,1\}$---following our methodology with the analogical transfer mechanism---enables the transformation of a multi-class classification problem involving approx\-imately $1,200$ distinct semantic roles in \fn{} into binary classification.

\subsection{Decoding Analogies}\label{sec:decoding}
Inspired by the notion of decoding 
termed 
in structured prediction approaches \cite{smith-2011}, we use the trained analogical model 
to obtain probability distributions over all semantic role candidates of a given frame. 
More precisely, given a target sentence of which the predicate $\mathtt{p^t}$ evokes frame $\phi$, as well as $k$ frame elements $\{\mathtt{e_i^t\}_{i=1}^k}$ identified, we view semantic role classification of each $\mathtt{e_i^t}$ as an independent event for which we compute probability distributions over all possible semantic roles types of $\phi$ by sampling a statistically significant $n_e$ number of source predicate-argument pairs $\{\mathtt{(p^s_j,e^s_j)}\}_{j=1}^{n_e}$ from source instances under the same frame type $\phi$.
The number $n_e$ remains identical to each and every semantic role that $\phi$ covers.
For each pair $\mathtt{(p^s_j,e^s_j)}$ and target $\mathtt{(p^t,e_i^t)}$ we apply our trained model $\mathcal{A}$. 
If $\mathcal{A}\mathtt{(p^s_j,e^s_j,p^t,e_i^t)}$ yields 1,
we apply analogical transfer, shortlisting the semantic role of $\mathtt{e^s_j}$ for $\mathtt{e_i^t}$. 
To obtain the final semantic role for target argument $\mathtt{e_i^t}$, we calculate the following score
for each semantic role $\rho_\kappa\in\phi$ with $\kappa$ representing the number of semantic role types in $\phi$:
\[ 
\sigma(\rho_\kappa) = \sum_{j=1}^{n_e}\mathcal{A}\mathtt{(p^s_j,e^s_j,p^t,e_i^t)} \text{ \small s.t. } sr(\phi, \mathtt{e^s_j}) = \rho_\kappa 
\]
Finally, we attribute to the target argument $\mathtt{e_i^t}$ the semantic role with the highest score:
\[
sr(\mathtt{e_i^t}) = \argmax_\kappa\left( \sigma(\rho_\kappa) \right) \]

\section{Experiments and Results}
\label{sec:expe}
In this section, we outline the experimental configuration and discuss the principal results.

\subsection{FrameNet Dataset}
For our experiments, we employ the latest \fn{} 1.7 dataset following the standard train-de\-ve\-lop\-ment-test partitions established in prior work of \citet{das-etal-2014-frame,Swayamdipta_etal_2017,graph:2021:lin-etal}.
This partitioning scheme has been consis\-tently adopted across the literature to ensure comparability with existing frame semantic parsing approaches. 
We extract all sentences from each partitioned document along with their associated annotations, organising them into corresponding training, development, and test splits---whose statistics are presented in Table~\ref{tab:data_retrieval}---for our experimental setup.
\begin{table}[thb]
    \centering
    \resizebox{0.8\linewidth}{!}{
    \begin{tabular}{lrrr}
        \toprule
        Dataset & N. sents   & N. frames  & N. docs \\
        \midrule
        All     & 27,228    & 796       & 101   \\
        train   & 18,772    & 754       & 70    \\
        dev     & 2,192     & 368       & 8     \\
        test    & 6,264     & 563       & 23    \\
        \bottomrule
    \end{tabular}}
    \caption{The number of sentences, frames, and documents in the datasets}
    \label{tab:data_retrieval}
\end{table}

\subsection{Analogy Dataset}
For analogical instance construction, we iterate within the scope of each semantic frame $\phi\in\Phi$ to extract all predicate-argument pairs present in the annotated sentences, yielding the frame-specific set $\mathtt{P_\phi}$. 
We subsequently generate the set $\mathtt{AP_\phi}$ of analogy instances, subsuming both negative and positive instances, through the Cartesian product
$\mathtt{AP_\phi} = \mathtt{P_\phi} \times \mathtt{P_\phi}$.
The complete set $\mathtt{AP_\Phi}$ of analogical instances is constructed as the union of all frame-specific sets $\mathtt{AP_\Phi} = \bigcup_{\phi\in\Phi} \mathtt{AP_\phi}$.

To mitigate class imbalance, we implemented dataset balancing techniques to create a balanced training partition \texttt{train\_balanced}, which serves as the training set for our experiments. 
Table~\ref{tab:ds_stats} presents the detailed distribution of instances across all data partitions.
\begin{table}[bth]
    \centering
    \resizebox{0.9\linewidth}{!}{
    \begin{tabular}{lrrr}
        \toprule
         Dataset                & all       & positive  & negative \\
         \midrule
         train                  & 7,638,128 & 3,036,470 & 4,601,658\\
         train\_balanced        & 6,072,940 & 3,036,470 & 3,036,470\\
         dev                    & 154,802   & 72,290    & 82,512\\
         test                   & 984,724   & 397,880   & 586,844\\
         \bottomrule
    \end{tabular}}
    \caption{Number of analogical instances}
    \label{tab:ds_stats}
\end{table}

\subsection{Analogical Model}

\begin{figure}[hbt]
    \centering
    \resizebox{0.55\linewidth}{!}{%
    \begin{tikzpicture}[
        node distance=0.8cm and 0.5cm,
        box/.style={rectangle, rounded corners=3pt, minimum width=3.2cm, minimum height=0.7cm, align=center, font=\sffamily, thick},
        encoder/.style={box, fill=cyan!15, draw=cyan!60!blue, minimum width=4.8cm},
        dense/.style={box, fill=orange!20, draw=orange!60!black},
        dropout/.style={box, fill=green!20, draw=green!60!black},
        decoder/.style={box, fill=yellow!20, draw=yellow!50!orange, minimum width=4.8cm},
        plus/.style={circle, draw=black, thick, fill=white, text=black, inner sep=1pt, minimum size=0.5cm, font=\bfseries\large},
        arrblack/.style={-{Stealth[scale=1]}, thick, draw=black!80},
        arrblue/.style={-{Stealth[scale=1]}, thick, draw=cyan!60!blue},
        arrorange/.style={-{Stealth[scale=1]}, thick, draw=orange!80!black},
        arrorangedashed/.style={arrorange, dashed},
        groupbox/.style={draw=black!70, rounded corners=12pt, thick, inner sep=8pt},
        dottedbox/.style={draw=black!60, densely dotted, thick, rounded corners=8pt, inner sep=8pt}
    ]

    \node[encoder] (encoder) {Encoder};

    \node[above=0.4cm of encoder.north, xshift=-1.8cm] (in_a) {$\langle{a}\rangle$};
    \node[above=0.4cm of encoder.north, xshift=-0.6cm] (in_b) {$\langle{b}\rangle$};
    \node[above=0.4cm of encoder.north, xshift=0.6cm]  (in_c) {$\langle{c}\rangle$};
    \node[above=0.4cm of encoder.north, xshift=1.8cm]  (in_d) {$\langle{d}\rangle$};

    \draw[arrblack] (in_a) -- (in_a |- encoder.north);
    \draw[arrblack] (in_b) -- (in_b |- encoder.north);
    \draw[arrblack] (in_c) -- (in_c |- encoder.north);
    \draw[arrblack] (in_d) -- (in_d |- encoder.north);

    \node[plus, below=0.5cm of encoder] (plus) {$+$};

    \draw[arrblue] ([xshift=-2cm]encoder.south) .. controls +(0,-0.6) and +(-0.6,0) .. (plus.west);
    \draw[arrblue] ([xshift=-0.7cm]encoder.south) .. controls +(0,-0.4) and +(0,0.5) .. ([xshift=-0.15cm]plus.north);
    \draw[arrblue] ([xshift=0.7cm]encoder.south) .. controls +(0,-0.4) and +(0,0.5) .. ([xshift=0.15cm]plus.north);
    \draw[arrblue] ([xshift=2cm]encoder.south) .. controls +(0,-0.6) and +(0.6,0) .. (plus.east);

    \node[dense, below=0.65cm of plus] (dense1) {Dense};
    \node[dropout, below=0.4cm of dense1] (dropout) {Dropout + Activ.};
    \node[dense, below=0.7cm of dropout] (dense2) {Dense};

    \draw[arrblue] (plus.south) -- (dense1.north);
    \draw[arrorange] (dense1.south) -- (dropout.north);
    \draw[arrorange] (dropout.south) -- (dense2.north);

    \node[dottedbox, fit=(dense1) (dropout)] (int_block) {};
    \node[anchor=north, font=\sffamily\scriptsize\color{black!60}, rotate=-90] (int_label) at ([xshift=0.08cm]int_block.east) {Intermediate block};
    
    \node[dottedbox, fit=(dense2)] (fin_block) {};
    \node[anchor=north, font=\sffamily\scriptsize\color{black!60}, rotate=-90] (fin_label) at ([xshift=0.08cm]fin_block.east) {Final block};

    \node[anchor=west, font=\sffamily\small\bfseries, rotate=-90, text=black!80] at ([xshift=0.1cm]int_label.north) {$*n$};

    \node[groupbox, fit=(int_block) (fin_block) (int_label) (fin_label)] (ana_model) {};
    \node[anchor=north, font=\sffamily\color{black!80}, rotate=-90] at ([xshift=0.5cm]ana_model.east) {Analogical Model};

    \node[decoder, below=0.9cm of dense2] (decoder) {Decoder};

    \draw[arrorangedashed] (dense2.south) -- (decoder.north);

    \end{tikzpicture}
    }
    \caption{Architecture of our pipeline}
    \label{fig:ff-model_arch}
\end{figure}
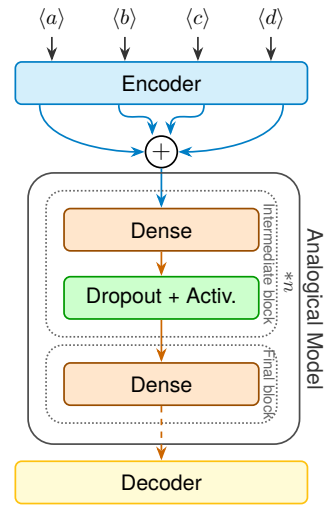

We design and train a simple feed-forward model, aiming to minimise the number of parameters required to solve the classification problem. 
The model architecture consists of a number of intermediate blocks followed by a final dense layer, where each intermediate block contains a dense, a dropout, and an activation layer. 
Throughout the feed-for\-ward network, the model learns weight matrices from concatenated input vectors and outputs a 2-dimensional vector representing class scores for NEG (invalid) and POS (valid).
We train this model of feed-forward architecture for binary classification of analogical relations. Hyperparameters are tuned to maximise accuracy on the development set. 
The hyperparameters tuned include: number of intermediate blocks, feature reduction rate function, dropout rate, and activation function for introducing non-linearity.
Experimental results demonstrate that a feed-forward model with two intermediate blocks, a geometric function
for feature reduction, a dropout rate of $0.3$, and ReLU activation functions achieves optimal training accuracy with only $\approx800,000$ parameters.
The model architecture is pictured 
in Figure~\ref{fig:ff-model_arch}.

\begin{figure}[bh]
    \centering
    \includegraphics[width=0.9\linewidth]{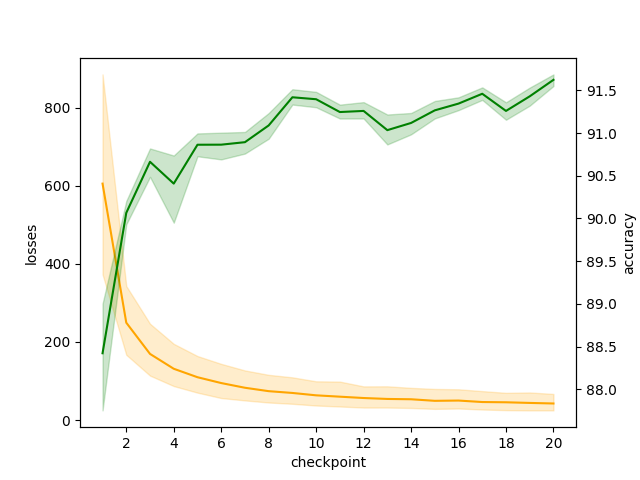}
    \caption{Training statistics wrt checkpoints}
    \label{fig:train-acc-loss}
\end{figure}

We train the model on embeddings of analogy instances obtained from the \texttt{bert-base-uncased} encoder~\cite{bert:naacl-hlt:2019:devlin-et-al}. 
Due to the substantial size of the dataset, we randomly partition the \texttt{train\_balanced} set into 20 disjoint data segments of similar sizes and employ incremental training, processing each segment sequentially until completion. 
Following each segment, we save the model to the corresponding checkpoint, resulting in a total of 20 checkpoints throughout the training process. 
The model optimises Cross-Entropy loss and converges rapidly to optimal performance within the initial segments, subsequently stabilising by the end of the approximately 7-minute training procedure on a single Tesla V100 GPU. 
Training accuracy and loss across checkpoints are shown in Figure~\ref{fig:train-acc-loss}.

We evaluate the binary analogical model on instances derived from the test partition and report per-class performance in Table~\ref{tab:eval:binary}. 
The model achieves strong results across both NEG and POS classes. 
A slight bias toward negative predictions is observed, which may be attributable to class imbalance in the development set used for validation.
\begin{table}[bt]
    \centering
    \resizebox{.80\linewidth}{!}{
        \begin{tabular}{lcc}
            \toprule
            Label  & 0 (negative) & 1(positive)\\
            \midrule
            Precision   & 91.80 & 86.03  \\
            Recall      & 90.18 & 88.23  \\
            F1 score    & 90.98 & 87.12  \\
            \midrule
            Accuracy    & \multicolumn{2}{c}{89.39}\\
            \bottomrule
        \end{tabular}
    }
    \caption{Scores of the binary analogy model (in \%)}
\label{tab:eval:binary}
\end{table}

\subsection{Analogical Transfer}
For every \textit{target} frame element $\mathtt{e_i^t}$ that needs classification in the test set, we extract the corresponding predicate $\mathtt{p^t}$ for the unique predicate-argument pair $\mathtt{(p^t,e_i^t)}$ of the \textit{target} set $\mathtt{P^t}$.
We subsequently build a source set $\mathtt{P^s_\rho}$ such that for each semantic role $\rho$ of the target frame type $\phi$, a fixed number $\mathit{n_{e}}$ of exemplary frame elements are retrieved from the \textit{source} dataset along with their corresponding predicates.
The statistics of our source dataset suggest $\mathit{n_{e}}=7$ following the minimum occurrence rate of $90\%$ of all semantic roles in this dataset.
We retrieve source predicate-arg\-ument pairs of form $\mathtt{(p_j^s,e_j^s)}$ such that:
\[
\forall \rho\in\phi, \mathtt{P_\rho^s} = \bigcup_{j=1}^{n_e}\{\mathtt{(p_j^s,e_j^s)}\} \text{ with } sr(\phi,e_j^s) = \rho 
\]
We perform classification using the trained model on the analogical instances in the set $\mathtt{AB_\rho}=\mathtt{P_t}\times\mathtt{P_\rho^s}$ and aggregate the scores of predicted positive instances to obtain the score for the corresponding role $\rho$.
The group with the highest score transfers its semantic role label to $\mathtt{e_i^t}$. 
Let us repeat that at no point during training nor inference was information on semantic roles passed to the analogical model. 

We evaluate the analogical transfer pipeline on the test set using the train set as the source of annotated sentences. 
This experimental design ensures that the model has never encountered the complete analogical instances---with target pairs from the test set---before evaluation, thereby maintaining the integrity of our experimental setup.

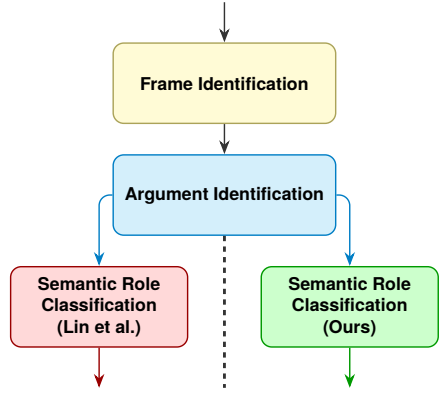
\begin{figure}[bt]
    \centering
    \resizebox{0.75\linewidth}{!}{
    \begin{tikzpicture}[
        node distance=0.8cm and 0.5cm,
        box/.style={rectangle, rounded corners=6pt, minimum width=4.4cm, minimum height=1.6cm, align=center, font=\sffamily\bfseries, thick},
        frameid/.style={box, fill=yellow!20, draw=yellow!60!black},
        argid/.style={box, fill=cyan!15, draw=cyan!60!blue},
        semrole1/.style={box, fill=red!15, draw=red!60!black, minimum width=3.5cm},
        semrole2/.style={box, fill=green!20, draw=green!60!black, minimum width=3.5cm},
        arrblack/.style={-{Stealth[scale=1.2]}, thick, draw=black!80},
        arrblue/.style={-{Stealth[scale=1.2]}, thick, draw=cyan!60!blue},
        arrred/.style={-{Stealth[scale=1.2]}, thick, draw=red!60!black},
        arrgreen/.style={-{Stealth[scale=1.2]}, thick, draw=green!60!black},
        dashedline/.style={dashed, ultra thick, draw=black!80}
    ]

    \node (start) {};
    
    \node[frameid, below=0.8cm of start] (frame) {Frame Identification};
    \draw[arrblack] (start) -- (frame.north);

    \node[argid, below=0.6cm of frame] (arg) {Argument Identification};
    \draw[arrblack] (frame.south) -- (arg.north);

    \node[semrole1, below left=0.6cm and -1.5cm of arg] (sem1) {Semantic Role\\Classification\\(Lin et al.)};
    \node[semrole2, below right=0.6cm and -1.5cm of arg] (sem2) {Semantic Role\\Classification\\(Ours)};

    \coordinate (mid_top) at ($(sem1.north)!0.5!(sem2.north)$);
    \coordinate (mid_bot) at ($(sem1.south)!0.5!(sem2.south)$);
    \draw[dashedline] (arg.south) -- (mid_top) -- ($(mid_bot) - (0,0.8cm)$);

    \draw[arrblue, rounded corners=6pt] (arg.west) -| (sem1.north);
    \draw[arrblue, rounded corners=6pt] (arg.east) -| (sem2.north);

    \node[below=0.8cm of sem1.south] (out1) {};
    \node[below=0.8cm of sem2.south] (out2) {};
    
    \draw[arrred] (sem1.south) -- (out1);
    \draw[arrgreen] (sem2.south) -- (out2);

    \end{tikzpicture}
    }
    \caption{Semantic parsing workflows with two distinct semantic role classifiers}
    \label{fig:semantic-parsing-pipeline}
\end{figure}

To compare our results with \sota{} approaches, we integrate our pipeline as a module for semantic role classification into the framework of \citet{graph:2021:lin-etal}, creating a complete end-to-end \fn{} Parsing system. 
This integration facilitates a direct comparison with existing methods by ensuring a level playing field. 
Figure~\ref{fig:semantic-parsing-pipeline} illustrates the experimental workflow and system architecture.

Table~\ref{tab:eval-srl} presents the comparative performance of both end-to-end semantic parsers. 
Our module as a semantic role classifier demonstrates improvements over the reported results, advancing the \sota{} on the \fn{} Parsing task. 
The improvements are attributable to our approach given the identical modules for Frame Identification and Argument Identification, proving the superior performance of our analogical transfer approach on the Semantic Role Classification task.
\begin{table}[bth]
    \centering
    \begin{tabular}{lllc}
        \toprule
        Frame id. & Arg id. & SR cl.      & Accuracy \\
        \midrule
        \citeauthor{graph:2021:lin-etal} & \citeauthor{graph:2021:lin-etal} & \citeauthor{graph:2021:lin-etal}  & 48.95 \\
        \citeauthor{graph:2021:lin-etal} & \citeauthor{graph:2021:lin-etal} & Ours                              & \textbf{49.81}\\
        \bottomrule
    \end{tabular}
    \caption{Overall performance of \sota{} end-to-end parser vs ours on \fn{} 1.7}
    \label{tab:eval-srl}
\end{table}

To further assess our analogical semantic role classifier, we integrated it into \texttt{Open-SE\-SA\-ME} ~\cite{Swayamdipta_etal_2017}
to perform classification on their predicted spans.
The experimental configuration, while similar to the framework presented in Figure~\ref{fig:semantic-parsing-pipeline}, employs gold-standard frame types and predicates as input.
We train their models on \fn{} 1.7 following their standard configuration.
Performance results for both approaches on the test set are presented in Table~\ref{tab:eval-sesame}, where we can observe again a clear performance gain of our approach. 

\begin{table}[bt]
    \centering
    \resizebox{\linewidth}{!}{
    \begin{tabular}{lllc}
        \toprule
        Frame id. & Arg id. & SR cl.      & Accuracy \\
        \midrule
        \texttt{gold} & \citeauthor{Swayamdipta_etal_2017} & \citeauthor{Swayamdipta_etal_2017}  & 58.46 \\
        \texttt{gold} & \citeauthor{Swayamdipta_etal_2017} & Ours                              & \textbf{59.35}\\
        \bottomrule
    \end{tabular}
    }
    \caption{Overall performance of former \sota{} SRL approach vs ours on \fn{} 1.7}
    \label{tab:eval-sesame}
\end{table}

Furthermore, to comprehensively evaluate our approach to semantic role classification, we assess its performance on instances derived from the test set using gold frame types and frame elements. 
The results are presented in Table~\ref{tab:eval_ongold}.
We include comparisons with two baselines, both employing similar MLP architecture 
for multi-class semantic role classification using \texttt{BERT} embeddings of the input. 
The first baseline performs classification on the sole frame element $\mathtt{e^t}$, while the second baseline combines predicates with frame elements $(\mathtt{p^t,e^t})$.
The results demonstrate that our analogy-based approach on predicate-frame element pairs $(\mathtt{p^s, e^s, p^t,e^t})$ substantially outperforms both baselines.

\begin{table}[hbt]
    \centering
    \resizebox{\linewidth}{!}{
        \begin{tabular}{lcccc}
            \toprule
              & Precision & Recall & F1 score & Accuracy \\
             \midrule
             Ours & 80.77 & 79.26 & 79.17 & 79.30 \\ 
             \midrule
             Baseline 1 ($\mathtt{e^t}$) & - & - & - & 9.14 \\
             Baseline 2 ($\mathtt{p^t, e^t}$) & - & - & - & 20.64 \\
             \bottomrule
        \end{tabular}
    }
    \caption{Baseline evaluation results on \fn{} 1.7 test set with gold frame and spans, using the exact same architecture as our approach. Precision, recall, and F1 scores are weighted.}
    \label{tab:eval_ongold}
\end{table}

\section{Related Work}\label{sec:rel_work}

\citet{Turney.08} introduced {Latent Relational Analysis} (LRA) for identifying analogies, which he tested in 20 scientific and metaphorical examples.
Later, \citet{efficient-representation-w2v:2013:mikolov,MikolovNIPS2013} 
initialised the trend of using analogies for word embedding quality evaluation, with Word2vec being the baseline encoder.
\citet{gladkova-etal-2016-analogy} demonstrated that poorly balanced datasets, such as the Google analogy test set \cite{MikolovNIPS2013,efficient-representation-w2v:2013:mikolov}, do not guarantee that static embeddings combined with a vector offset approach are able to capture analogies.
This led the authors to introduce the Bigger Analogy Test Set (BATS), showing that 
even with balanced data,
derivational and lexicographic relations remain a challenge.
\citet{rogers-etal-2017-many} showed that the vector offset approach as well as 3CosAdd \cite{levy-goldberg-2014-linguistic} suffer from dependence on vector similarity, thus arguing against the use of such datasets to evaluate the intrinsic 
quality
of word embeddings.

\citet{sultan-shahaf-2022-life} adapted Structure Mapping Theory \cite{Gentner.83} to procedural texts, extracting entities and their relationships, and finding suitable mappings between two different domains based on relational similarity. 
Relations are sets of ordered verbs between entities that are extracted on the basis of question/answer pairs. 
Here, higher similarity suggests that sets share more relations. 
Mappings are identified heuristically based on the cosine similarity of \texttt{BERT} vectors representing the questions that provided the entities. 
Their approach successfully extracts mappings in two different datasets.

On Semantic Role Classification, 
\citet{gildea2002automatic} introduced a statistical classifier to classify spans into corresponding semantic roles on \fn{}. 
Their approach requires a deal of rigorous processing and laborious manual annotation of the sentences that engineer the features for classification. 
The authors reported 80.4\% accuracy on gold spans.
Comparing that with our results of 80.62\% accuracy on gold spans is not straightforward due to the fact that (i) we do not work on the same version of \fn{}---we use  \fn{} 1.7 while they \fn{} 1.5---as well as (ii) the two datasets exhibit statistical differences in the train--dev--test splits
such as the number of spans; we classified 11,206 spans while they did 8,167.
Similar feature engineering approaches were also adopted by \citet{cohn2005semantic} to tackle the SRL problem on the CoNLL shared task and dataset \cite{carreras2005introduction}.
Syntactic parsing -- as a technique to obtain 
syntactic
features from structured texts -- is deemed one of the key factors in their success, 
consistent with the finding of
\citet{wang2019best}.

To the best of our knowledge, no existing work has ever employed analogies to solve similar problems in \fn{} Parsing and its sub-tasks.
\citet{Swayamdipta_etal_2017} presented a Softmax-margin semi-Markov model. 
Their 
\texttt{Open-SESAME} framework
initially combines a bidirectional RNN with \texttt{SegRNN} \cite{gimpel2010softmax} 
then employs syntactic scaffolding to further its initial results to \sota.
More recently, \citet{graph:2021:lin-etal} use Graph Neural Networks 
based on \texttt{BERT} embeddings and \texttt{BiHLSTM} architecture \cite{Srivastava_etal_NIPS_2015} for 
full \fn{} Parsing, achieving new highs.
Finally, \citet{chanin2023opensource} implemented a framework that fine-tunes a \texttt{T5} model \cite{Raffel_etal:t5} on \fn{} and PropBank \cite{Kingsbury2002} data, rivalling the above parsers.

\section{Discussion and Perspectives}
\label{sec:disc}
Following concerns over coverage gaps in semantic role labelling addressed by \citet{palmer2010evaluating}, we examined our results in order to gain valuable insights and identify potential improvements to our methodology.
Among the five gaps discussed by \citet{palmer2010evaluating}, we hypothesise that our model may be susceptible to no-training-data (NOTR) gaps, attributable to the differences between frame types and semantic roles covered across the train, development, and test partitions.
When evaluated on approximately $1{,}800$ in\-stan\-ces across $238$ out of $563$ frame types present in the test set but entirely absent from the development set, which makes up more than $42\%$ of total frame types evaluated, the model suffered a $\approx 4\%$ accuracy loss.
On approximately $2{,}000$ test instances containing semantic roles absent from the development set, performance loss increased to $\approx7\%$.
While these results present losses compared to the overall accuracy of $79.30\%$ on test set, the comparably acceptable rates demonstrate the model's generalisation capability across unseen properties in line with the findings of \citet{afantenos:learning}.
Though not strictly impactful, these limitations may be mitigated by enriching the training and development sets with additional resources such as CoNLL-05 \cite{carreras2005introduction} and CoNLL-09 \cite{hajic2009conll} datasets as suggested in \citet{fitzgerald2015semantic}.

On the results of recent approaches, namely \texttt{Framenet-Parser}~\cite{graph:2021:lin-etal} and \texttt{Open-SESAME}~\cite{Swayamdipta_etal_2017}, we encountered discrepancies between the reported and reproduced scores. 
On \texttt{Framenet-Parser}, though we used the reported score of $48.95\%$ for comparison, the reproduced accuracy of $47.85\%$ reflected original inaccuracies in both argument identification and classification tasks, consequently having a negative effect on our classifier that relies on their argument identification module.
Their SRL module also remains inaccessible, hindering the possibilities for separate module evaluation.
For \texttt{Open-SESAME}, the inconsistency seems consistent across \fn{} 1.5 and 1.7 datasets.
The authors acknowledged the problem on their repository\footnote{\href{https://github.com/swabhs/open-sesame}{https://github.com/swabhs/open-sesame}}.
Since their pre-trained \texttt{argid} model on \fn{} 1.7 cannot be loaded directly due to configuration mismatch, we had to reproduce their standalone model from scratch and reported on the collected results.
That being said, even the improvement by our pipeline remains consistent given the best effort we invested in reproducibility across experiments, further investigation is encouraged to discover any overlooked variables.

Our experiments were designed and conducted around \fn{} with proof-of-concept objectives rather than providing a holistic approach to semantic parsing or semantic role labelling tasks.
Beyond the performance gains, this work is designated to introduce a novel direction for knowledge and representation transfer that transforms multi-label classification into intuitive, human-interpretable, explainable, robust, and scalable binary classification.
The reasoning process relies entirely on human-like mechanisms that enhance transparency and explainability.
Unlike traditional classification approaches, our analogical model requires no retraining when new labels emerge, highlighting the competitive scalability of the analogical transfer approach.

Regarding computation and energy efficiency, the entire pipeline can be trained and evaluated on a single Tesla V100 GPU---including \texttt{BERT} embedding extraction---within 15 minutes.
Consuming approximately 50Wh, our approach achieves \sota{} performance while remaining considerably more energy-efficient than existing methods.

In this study, we selected \texttt{bert-base-uncased} based on the model's computational efficiency and simplicity rather than optimal performance.
Future work will examine top notch models from the \textsc{MTEB} leaderboard\footnote{\href{https://huggingface.co/spaces/mteb/leaderboard}{https://huggingface.co/spaces/mteb/leaderboard}}~\cite{muennighoff2022mteb}, as well as non-contextualised encoders such as GloVe~\cite{pennington-etal-2014-glove}.
This is necessary to address and assess the correlation between the representation quality and downstream model performance.

Our additional experiments also indicate that sentence embeddings can also serve as representations of context, albeit with a modest decrease in overall performance (approximately $4\%$ in this \fn{} Parsing task). 
Nevertheless, this suggests promising directions, as it allows for more flexible constructions of analogy instances and may open new avenues for solving other problems, especially when encapsulating the context through its components is infeasible.
That being said, we plan to evaluate the analogical model on other tasks beyond Semantic Role Classification to explore the broader applicability of analogical transfer.
We also foresee the expansion of the analogical transfer methodology to and across other datasets to learn whether comparable performance improvements can be achieved while preserving parameter efficiency and model frugality.

\section{Conclusion}\label{sec:conclusion}
In this paper, we adopted a relational view of analogies that allows us to transform a multi-class classification task into binary classification, while recuperating original classes during inference by decoding model output into probability distributions with random sampling. 
We applied our approach to the Semantic Role Classification task on \fn{} 1.7. 
We define two binary relations, representing valid and invalid instances of analogies, using pairs of 
frame evoking predicates
and frame elements covered by the same semantic frame $\phi$ as the domain and co-domain of both relations. 
Frame elements of valid instances share the same semantic role type, while invalid ones do not. 

Using \texttt{BERT} embeddings to train a lightweight ANN with minimal parameters, we have achieved state-of-the-art performance on this task.
Our approach exhibits strong energy efficiency, with rapid convergence suggesting potential for further optimisation.
The absence of explicit class information during training enables analogical transfer to new classes without model retraining, requiring only annotated examples for sampling during inference, thereby enhancing both scalability and energy efficiency. 
At the same time, our method preserves interpretability through analogical transfer, providing transparent reasoning mechanisms that promote human understanding of model decisions.
We publish the detailed implementation of this work on our repository\footnote{https://github.com/thebugcreator/franabelling}.


\appendix

\section*{Limitations}
Optimal decoding of the binary model's output assumes a sufficiently large number of annotated instances for reference. 
Failure to obtain necessary reference instances may hinder accurate analogical transfer of under-represented classes
at the decoding phase. 
Nonetheless, 
this not-enough-data gap is a rather general problem in Machine Learning \cite{Ghosh2024}.


\section*{Acknowledgements}
This work is supported by the AT2TA project\footnote{https://at2ta.loria.fr/} funded by the French National Research Agency (``Agence Nationale de la Recherche'' – ANR) under grant ANR-22-CE23-0023. 

\noindent This work has also been supported by the Occitanie region, the European Regional Development Fund (ERDF), and the French government through the France 2030 project managed by the ANR with reference number ANR-22-EXES-0015.

\section*{References}
\bibliographystyle{lrec2026-natbib}
\bibliography{analogies}

@inproceedings{CunhaAAAI2026,
  author       = {Francisco Cunha and
                Yves Lepage and 
                Miguel Couceiro and
                Zied Bouraoui},
  title        = {Generalizing Analogical Inference from Boolean to Continuous Domains},
booktitle    = {AAAI-26, Sponsored by the Association for the Advancement of Artificial                  Intelligence, January 20 - January 27, 2026, Singapore},
  pages        = {},
  publisher    = {{AAAI} Press. To appear},
  year         = {2026},
  url = {https://doi.org/10.48448/ncay-r562} 
  }

@article{Ghosh2024,
  author  = {Ghosh, Kushankur and Bellinger, Colin and Corizzo, Roberto and Branco, Paula and Krawczyk, Bartosz and Japkowicz, Nathalie},
  title   = {The class imbalance problem in deep learning},
  journal = {Machine Learning},
  year    = {2024},
  volume  = {113},
  number  = {7},
  pages   = {4845--4901},
  month   = {7},
  doi     = {10.1007/s10994-022-06268-8},
  url     = {https://doi.org/10.1007/s10994-022-06268-8},
  issn    = {1573-0565}
}

@book{firth1968selected,
  title        = {Selected Papers of {J. R. Firth}, 1952--59},
  author       = {Firth, John Rupert},
  editor       = {Palmer, Frank Robert},
  year         = {1968},
  publisher    = {Indiana University Press},
  address      = {Bloomington},
  series       = {Indiana University Studies in the History and Theory of Linguistics}
}

@article{harris1954distributional,
  title        = {Distributional Structure},
  author       = {Harris, Zellig S.},
  journal      = {Word},
  volume       = {10},
  number       = {2--3},
  pages        = {146--162},
  year         = {1954},
  publisher    = {Routledge},
  doi          = {10.1080/00437956.1954.11659520}
}

@article{MarquerC25,
  author       = {Esteban Marquer and
                  Miguel Couceiro},
  title        = {Solving morphological analogies: from retrieval to generation},
  journal      = {Ann. Math. Artif. Intell.},
  volume       = {93},
  number       = {2},
  pages        = {263--298},
  year         = {2025},
  bibsource    = {dblp computer science bibliography, https://dblp.org}
}

@article{Bounhas.Prade.2023,
title = {Analogy-based classifiers: An improved algorithm exploiting competent data pairs},
journal = {International Journal of Approximate Reasoning},
volume = {158},
pages = {108923},
year = {2023},
issn = {0888-613X},
author = {Myriam Bounhas and Henri Prade}
}

@article{chanin2023opensource,
  title={Open-source frame semantic parsing},
  author={Chanin, David},
  journal={arXiv preprint arXiv:2303.12788},
  year={2023}
}

@article{Firt:analogical:2025,
  title={Analogical reasoning as a core AGI capability},
  author={Firt, Erez},
  journal={AI and Ethics},
  volume={5},
  number={5},
  pages={5501--5515},
  year={2025},
  publisher={Springer}
}

@inproceedings{pennington-etal-2014-glove,
    title = "{G}lo{V}e: Global Vectors for Word Representation",
    author = "Pennington, Jeffrey  and
      Socher, Richard  and
      Manning, Christopher",
    editor = "Moschitti, Alessandro  and
      Pang, Bo  and
      Daelemans, Walter",
    booktitle = "Proceedings of the 2014 Conference on Empirical Methods in Natural Language Processing ({EMNLP})",
    month = oct,
    year = "2014",
    address = "Doha, Qatar",
    publisher = "Association for Computational Linguistics",
    pages = "1532--1543"
}

@book{smith-2011,
	author = {Noah A. Smith},
	month = {May},
	publisher = {Morgan and Claypool},
	series = {Synthesis Lectures on Human Language Technologies},
	title = {Linguistic Structure Prediction},
	year = {2011}}

@article{Raffel_etal:t5,
  title={Exploring the limits of transfer learning with a unified text-to-text transformer},
  author={Raffel, Colin and Shazeer, Noam and Roberts, Adam and Lee, Katherine and Narang, Sharan and Matena, Michael and Zhou, Yanqi and Li, Wei and Liu, Peter J},
  journal={The Journal of Machine Learning Research},
  volume={21},
  number={1},
  pages={5485--5551},
  year={2020},
  publisher={JMLRORG}
}

@inproceedings{Kingsbury2002,
author = {Kingsbury, Paul and Palmer, Martha},
booktitle = {Proceedings of the 3rd International Conference on Language Resources and Evaluation (LREC-2002), Las Palmas, Spain},
title = {{From Treebank to PropBank}},
year = {2002}
}

@inproceedings{graph:2021:lin-etal,
    title = "A Graph-Based Neural Model for End-to-End Frame Semantic Parsing",
    author = "Lin, ZhiChao  and
      Sun, Yueheng  and
      Zhang, Meishan",
    booktitle = "Conference on Conference on Empirical Methods in Natural Language Processing",
    month = nov,
    year = "2021",
    address = "Online and Punta Cana, Dominican Republic",
    publisher = "Association for Computational Linguistics",
    pages = "3864--3874",
}

@article{barbot_analogy_2019,
	title = {Analogy between concepts},
	volume = {275},
	pages = {487--539},
	journal = {Artificial Intelligence},
	author = {Barbot, Nelly and Miclet, Laurent and Prade, Henri},
	year = {2019},
}

@inproceedings{DBLP:conf/iccbr/PradeR17,
  author       = {Henri Prade and
                  Gilles Richard},
  title        = {Analogical Proportions and Analogical Reasoning - An Introduction},
  booktitle    = {Case-Based Reasoning Research and Development - 25th International
                  Conference, {ICCBR} 2017, Trondheim, Norway, June 26-28, 2017, Proceedings},
  series       = {LNCS},
  volume       = {10339},
  pages        = {16--32},
  publisher    = {Springer},
  year         = {2017}
}

@book{hofstadter_surfaces_2013,
	title = {Surfaces and Essences: Analogy as the Fuel and Fire of Thinking},
	publisher = {Basic Books},
	author = {Hofstadter, Douglas R. and Sander, Emmanuel},
	year = {2013},
}

@article{chalmers_high-level_1992,
  title={High-level perception, representation, and analogy: A critique of artificial intelligence methodology},
  author={Chalmers, David J. and French, Robert M. and Hofstadter, Douglas R.},
  journal={Journal of Experimental \& Theoretical Artificial Intelligence},
  volume={4},
  number={3},
  pages={185--211},
  year={1992},
  publisher={Taylor \& Francis}
}

@report{hofstadter_copycat_1984,
  title={The Copycat Project: An Experiment in Nondeterminism and Creative Analogies.},
  author={Hofstadter, Douglas R.},
  year={1984}
}

@inproceedings{ushio-etal-2021-bert,
    title = "{BERT} is to {NLP} what {A}lex{N}et is to {CV}: Can Pre-Trained Language Models Identify Analogies?",
    author = "Ushio, Asahi  and
      Espinosa Anke, Luis  and
      Schockaert, Steven  and
      Camacho-Collados, Jose",
    editor = "Zong, Chengqing  and
      Xia, Fei  and
      Li, Wenjie  and
      Navigli, Roberto",
    booktitle = "Proceedings of the 59th Annual Meeting of the Association for Computational Linguistics and the 11th International Joint Conference on Natural Language Processing (Volume 1: Long Papers)",
    month = aug,
    year = "2021",
    address = "Online",
    publisher = "Association for Computational Linguistics",
    pages = "3609--3624",
}

@inproceedings{lepage_translation_2005,
	location = {Poznań, Poland},
	title = {Translation of Sentences by Analogy Principle},
	booktitle = {Second Language and Technology Conference},
	author = {Lepage, Yves},
	year = {2005},
}

@misc{petersen_et_al:2025,
      title={Modelling Analogies and Analogical Reasoning: Connecting Cognitive Science Theory and NLP Research}, 
      author={Molly R. Petersen and Claire. E Stevenson and Lonneke van der Plas},
      year={2025},
      archivePrefix={arXiv},
      primaryClass={cs.CL},
}

@article{gentner_hoyos:2017,
  author  = {Gentner, Dedre and Hoyos, Christian},
  title   = {Analogy and Abstraction},
  journal = {Topics in Cognitive Science},
  volume  = {9},
  number  = {3},
  pages   = {672--693},
  year    = {2017},
  publisher = {Wiley},
}

@inproceedings{petersen-van-der-plas-2023-language,
    title = "Can language models learn analogical reasoning? Investigating training objectives and comparisons to human performance",
    author = "Petersen, Molly  and
      van der Plas, Lonneke",
    editor = "Bouamor, Houda  and
      Pino, Juan  and
      Bali, Kalika",
    booktitle = "Proceedings of the 2023 Conference on Empirical Methods in Natural Language Processing",
    month = dec,
    year = "2023",
    address = "Singapore",
    publisher = "Association for Computational Linguistics",
    pages = "16414--16425",
}

@inproceedings{ulcar-etal-2020-multilingual,
    title = "Multilingual Culture-Independent Word Analogy Datasets",
    author = {Ul{\v{c}}ar, Matej  and
      Vaik, Kristiina  and
      Lindstr{\"o}m, Jessica  and
      Dailid{\.{e}}nait{\.{e}}, Milda  and
      Robnik-{\v{S}}ikonja, Marko},
    editor = "Calzolari, Nicoletta  and
      B{\'e}chet, Fr{\'e}d{\'e}ric  and
      Blache, Philippe  and
      Choukri, Khalid  and
      Cieri, Christopher  and
      Declerck, Thierry  and
      Goggi, Sara  and
      Isahara, Hitoshi  and
      Maegaard, Bente  and
      Mariani, Joseph  and
      Mazo, H{\'e}l{\`e}ne  and
      Moreno, Asuncion  and
      Odijk, Jan  and
      Piperidis, Stelios",
    booktitle = "Proceedings of the Twelfth Language Resources and Evaluation Conference",
    month = may,
    year = "2020",
    address = "Marseille, France",
    publisher = "European Language Resources Association",
    pages = "4074--4080",
    language = "eng",
    ISBN = "979-10-95546-34-4",
}

@inproceedings{karpinska-etal-2018-subcharacter,
    title = "Subcharacter Information in {J}apanese Embeddings: When Is It Worth It?",
    author = "Karpinska, Marzena  and
      Li, Bofang  and
      Rogers, Anna  and
      Drozd, Aleksandr",
    editor = "Dinu, Georgiana  and
      Ballesteros, Miguel  and
      Sil, Avirup  and
      Bowman, Sam  and
      Hamza, Wael  and
      Sogaard, Anders  and
      Naseem, Tahira  and
      Goldberg, Yoav",
    booktitle = "Proceedings of the Workshop on the Relevance of Linguistic Structure in Neural Architectures for {NLP}",
    month = jul,
    year = "2018",
    address = "Melbourne, Australia",
    publisher = "Association for Computational Linguistics",
    pages = "28--37",
}

@book{hofstadter_fluid_1995,
	location = {New York},
	title = {Fluid Concepts and Creative Analogies: Computer Models of the Fundamental Mechanisms of Thought},
	shorttitle = {Fluid Concepts and Creative Analogies},
	publisher = {Basic Books},
	author = {Hofstadter, Douglas R. and {FARG}},
	year = {1995},
	note = {Publication Title: Fluid Concepts and Creative Analogies: Computer Models of the Fundamental Mechanisms of Thought},
}

@book{holyoak_mental_1995,
	location = {Cambridge, Massachusetts},
	title = {Mental Leaps: Analogy in Creative Thought},
	series = {A Bradford Book},
	publisher = {{MIT} Press},
	author = {Holyoak, Keith J. and Thagard, Paul},
	year = {1995},
}

@book{mitchell_analogy_1993,
	location = {Cambridge, Massachusetts},
	title = {Analogy Making as Perception: A Computer Model},
	series = {Neural Network Modeling and Connectionism},
	publisher = {The {MIT} Press},
	author = {Mitchell, Melanie},
	year = {1993},
}

@inproceedings{prade2009analogy,
  title={Analogy, paralogy and reverse analogy: Postulates and inferences},
  author={Prade, Henri and Richard, Gilles},
  booktitle={Annual conference on artificial intelligence},
  pages={306--314},
  year={2009},
  organization={Springer}
}

@inproceedings{bert:naacl-hlt:2019:devlin-et-al,
  author    = {Jacob Devlin and
               Ming{-}Wei Chang and
               Kenton Lee and
               Kristina Toutanova},
  title     = {{BERT:} Pre-training of Deep Bidirectional Transformers for Language
               Understanding},
  booktitle = {North American Chapter of the Association for Computational Linguistics: Human Language Technologies},
Volume = {1},
  pages     = {4171--4186},
  publisher = {Association for Computational Linguistics},
  year      = {2019},
}

@incollection{Hofstadter.01,
	address = {Cambridge, Massachusetts},
	author = {Douglas R. Hofstadter},
	booktitle = {{The Analogical Mind: Perspectives from Cognitive Science}},
	chapter = 15,
	pages = {499--538},
	publisher = {The MIT Press},
	title = {{Analogy as the Core of Cognition}},
	year = 2001,
 editor = {Dedre Gentner and Keith J. Holyoak and Boicho N. Kokinov},
}

@article{Turney.08,
	author = {Peter D. Turney},
	journal = {Journal of Artificial Intelligence Research},
	pages = {615--655},
	title = {{The Latent Relation Mapping Engine: Algorithm and Experiments}},
	volume = 33,
	year = 2008,
 }

@inproceedings{efficient-representation-w2v:2013:mikolov,
    author    = {Tom{\'{a}}s Mikolov and Kai Chen and
               Greg Corrado and
               Jeffrey Dean},
  title     = {Efficient Estimation of Word Representations in Vector Space},
  booktitle = {International Conference on Learning Representations, Workshop},
  year      = {2013},
}

@incollection{MikolovNIPS2013,
	author = {Tom{\'{a}}s Mikolov and
               Ilya Sutskever and
               Kai Chen and
               Gregory S. Corrado and
               Jeffrey Dean},
	booktitle = {Advances in Neural Information Processing Systems},
    volume = {26},
	pages = {3111--3119},
	publisher = {Curran Associates Inc.},
	title = {Distributed Representations of Words and Phrases and their Compositionality},
	year = {2013},
 }

@inproceedings{gladkova-etal-2016-analogy,
    title = "Analogy-based detection of morphological and semantic relations with word embeddings: what works and what doesn{'}t.",
    author = "Gladkova, Anna  and
      Drozd, Aleksandr  and
      Matsuoka, Satoshi",
    booktitle = "North American Chapter of the Association for Computational Linguistics, Student Research Workshop",
    year = "2016",
    address = "San Diego, California",
    publisher = "Association for Computational Linguistics",
    pages = "8--15",
}

@inproceedings{rogers-etal-2017-many,
    title = "The (too Many) Problems of Analogical Reasoning with Word Vectors",
    author = "Rogers, Anna  and
      Drozd, Aleksandr  and
      Li, Bofang",
    booktitle = "Joint Conference on Lexical and Computational Semantics",
    year = "2017",
    address = "Vancouver, Canada",
    publisher = "Association for Computational Linguistics",
    pages = "135--148",
}

@inproceedings{baker-etal-1998-berkeley-framenet,
    title = "The {B}erkeley {F}rame{N}et Project",
    author = "Baker, Collin F.  and
      Fillmore, Charles J.  and
      Lowe, John B.",
    booktitle = "International Conference on Computational Linguistics",
    Volume  ="1",
    month = aug,
    year = "1998",
    address = "Montreal, Quebec, Canada",
    publisher = "Association for Computational Linguistics",
    pages = "86--90",
}

@inproceedings{levy-goldberg-2014-linguistic,
    title = "Linguistic Regularities in Sparse and Explicit Word Representations",
    author = "Levy, Omer  and
      Goldberg, Yoav",
    booktitle = "Proceedings of the Eighteenth Conference on Computational Natural Language Learning",
    month = jun,
    year = "2014",
    address = "Ann Arbor, Michigan",
    publisher = "Association for Computational Linguistics",
    pages = "171--180",
}

@misc{Swayamdipta_etal_2017,
      title={Frame-Semantic Parsing with Softmax-Margin Segmental RNNs and a Syntactic Scaffold}, 
      author={Swabha Swayamdipta and Sam Thomson and Chris Dyer and Noah A. Smith},
      year={2017},
      archivePrefix={arXiv},
      primaryClass={cs.CL},
}

@article{bakerarticle2017framenet,
  title={Framenet: Frame semantic annotation in practice},
  author={Baker, Collin F.},
  journal={Handbook of Linguistic Annotation},
  pages={771--811},
  year={2017},
  publisher={Springer},
}

@inproceedings{baker-etal-2007-semeval,
    title = "{S}em{E}val-2007 Task 19: Frame Semantic Structure Extraction",
    author = "Baker, Collin F.  and
      Ellsworth, Michael  and
      Erk, Katrin",
    editor = "Agirre, Eneko  and
      M{\`a}rquez, Llu{\'i}s  and
      Wicentowski, Richard",
    booktitle = "Proceedings of the Fourth International Workshop on Semantic Evaluations ({S}em{E}val-2007)",
    month = jun,
    year = "2007",
    address = "Prague, Czech Republic",
    publisher = "Association for Computational Linguistics",
    pages = "99--104"
}

@inproceedings{drozd-etal-2016-word,
    title = "Word Embeddings, Analogies, and Machine Learning: Beyond king - man + woman = queen",
    author = "Drozd, Aleksandr  and
      Gladkova, Anna  and
      Matsuoka, Satoshi",
    booktitle = "International Conference on Computational Linguistics, Technical Papers",
    year = "2016",
    address = "Osaka, Japan",
    publisher = "The COLING 2016 Organizing Committee",
    pages = "3519--3530",
}

@inproceedings{mickus-etal-2023-mann,
    title = "{\quotedblbase}Mann{\textquotedblleft} is to {\textquotedblleft}Donna{\textquotedblright} as  {\textquotedblleft}gu\'ow\'ang{\textquotedblright} is to {\guillemotleft}Reine{\guillemotright} Adapting the Analogy Task for Multilingual and Contextual Embeddings",
    author = "Mickus, Timothee  and
      Cal{\`o}, Eduardo  and
      Jacqmin, L{\'e}o  and
      Paperno, Denis  and
      Constant, Mathieu",
    editor = "Palmer, Alexis  and
      Camacho-collados, Jose",
    booktitle = "Proceedings of the 12th Joint Conference on Lexical and Computational Semantics (*SEM 2023)",
    month = jul,
    year = "2023",
    address = "Toronto, Canada",
    publisher = "Association for Computational Linguistics",
    pages = "270--283"
}

@book{ruppenhofer2006extended,
	address = {Berkeley, California},
	author = {Ruppenhofer, Josef and Ellsworth, Michael and Petruck, Miriam R.L. and Johnson, Christopher R. and Scheffczyk, Jan},
	note = {Distributed with the FrameNet data},
	organization = {International Computer Science Institute},
	publisher = {International Computer Science Institute},
	title = {{FrameNet II: Extended Theory and Practice}},
	year = {2006}
}

@inproceedings{Srivastava_etal_NIPS_2015,
 author = {Srivastava, Rupesh K. and Greff, Klaus and Schmidhuber, J\"{u}rgen},
 booktitle = {Advances in Neural Information Processing Systems},
 pages = {},
 publisher = {Curran Associates, Inc.},
 title = {Training Very Deep Networks},
 volume = {28},
 year = {2015}
}

@inproceedings{sultan-shahaf-2022-life,
    title = "Life is a Circus and We are the Clowns: Automatically Finding Analogies between Situations and Processes",
    author = "Sultan, Oren  and
      Shahaf, Dafna",
    booktitle = "Conference on Empirical Methods in Natural Language Processing",
    month = dec,
    year = "2022",
    address = "Abu Dhabi, United Arab Emirates",
    publisher = "Association for Computational Linguistics",
    pages = "3547--3562",
}

@article{Gentner.83,
	author = {Dedre Gentner},
	journal = {Cognitive Science},
	pages = {155--170},
	title = {{Structure Mapping: A Theoretical Framework for Analogy}},
	volume = 7,
	year = 1983,
  }

@inproceedings{zheng-etal:2022,
    title = "A Double-Graph Based Framework for Frame Semantic Parsing",
    author = "Zheng, Ce  and
      Chen, Xudong  and
      Xu, Runxin  and
      Chang, Baobao",
    editor = "Carpuat, Marine  and
      de Marneffe, Marie-Catherine  and
      Meza Ruiz, Ivan Vladimir",
    booktitle = "Proceedings of the 2022 Conference of the North American Chapter of the Association for Computational Linguistics: Human Language Technologies",
    month = jul,
    year = "2022",
    address = "Seattle, United States",
    publisher = "Association for Computational Linguistics",
    pages = "4998--5011",
}

@article{das-etal-2014-frame,
    title = "Frame-Semantic Parsing",
    author = "Das, Dipanjan  and
      Chen, Desai  and
      Martins, Andr{\'e} F. T.  and
      Schneider, Nathan  and
      Smith, Noah A.",
    journal = "Computational Linguistics",
    volume = "40",
    number = "1",
    month = mar,
    year = "2014",
    address = "Cambridge, MA",
    publisher = "MIT Press",

    pages = "9--56",
}

@article{gildea2002automatic,
  title={Automatic labeling of semantic roles},
  author={Gildea, Daniel and Jurafsky, Daniel},
  journal={Computational linguistics},
  volume={28},
  number={3},
  pages={245--288},
  year={2002},
  publisher={MIT Press One Rogers Street, Cambridge, MA 02142-1209, USA journals-info~…}
}

@inproceedings{cohn2005semantic,
  title={Semantic role labelling with tree conditional random fields},
  author={Cohn, Trevor and Blunsom, Phil},
  booktitle={Proceedings of the Ninth Conference on Computational Natural Language Learning (CoNLL-2005)},
  pages={169--172},
  year={2005}
}

@inproceedings{carreras2005introduction,
  title={Introduction to the CoNLL-2005 shared task: Semantic role labeling},
  author={Carreras, Xavier and M{\`a}rquez, Llu{\'\i}s},
  booktitle={Proceedings of the ninth conference on computational natural language learning (CoNLL-2005)},
  pages={152--164},
  year={2005}
}

@article{wang2019best,
  title={How to best use syntax in semantic role labelling},
  author={Wang, Yufei and Johnson, Mark and Wan, Stephen and Sun, Yifang and Wang, Wei},
  journal={arXiv preprint arXiv:1906.00266},
  year={2019}
}

@inproceedings{palmer2010evaluating,
  title={Evaluating FrameNet-style semantic parsing: the role of coverage gaps in FrameNet},
  author={Palmer, Alexis and Sporleder, Caroline},
  booktitle={Coling 2010: Posters},
  pages={928--936},
  year={2010}
}

@inproceedings{fitzgerald2015semantic,
  title={Semantic role labeling with neural network factors},
  author={FitzGerald, Nicholas and T{\"a}ckstr{\"o}m, Oscar and Ganchev, Kuzman and Das, Dipanjan},
  booktitle={Proceedings of the 2015 Conference on Empirical Methods in Natural Language Processing},
  pages={960--970},
  year={2015}
}

@inproceedings{hajic2009conll,
  title={The CoNLL-2009 shared task: Syntactic and semantic dependencies in multiple languages},
  author={Hajic, Jan and Ciaramita, Massimiliano and Johansson, Richard and Kawahara, Daisuke and Mart{\'\i}, M Ant{\`o}nia and M{\`a}rquez, Llu{\'\i}s and Meyers, Adam and Nivre, Joakim and Pad{\'o}, Sebastian and {\v{S}}t{\v{e}}p{\'a}nek, Jan and others},
  booktitle={Proceedings of the Thirteenth Conference on Computational Natural Language Learning (CoNLL 2009): Shared Task},
  pages={1--18},
  year={2009}
}

@article{muennighoff2022mteb,
  title={Mteb: Massive text embedding benchmark},
  author={Muennighoff, Niklas and Tazi, Nouamane and Magne, Lo{\"\i}c and Reimers, Nils},
  journal={arXiv preprint arXiv:2210.07316},
  year={2022}
}

@inproceedings{ afantenos:learning,
  author = "Stergos Afantenos and Miguel Couceiro and Emiliano Lorini and Van-Duy Ngo",
  title = "Learning Proportional Analogies: Lightweight Neural Network vs Large Language Models",
  booktitle = "Proceedings of the 18th International Conference on Agents and Artificial Intelligence",
  pages = "3604--3614",
  year = "2026",
  isbn = "978-989-758-796-2",
  issn = "2184-433X"
}

@book{Aristotle2009,
  author       = {Aristotle},
  title        = {Nicomachean Ethics},
  translator   = {D. Ross},
  publisher    = {Oxford University Press},
  series       = {Oxford World's Classics},
  address      = {Oxford},
  year         = {2009}
}

@article{fillmore1976frame,
  title={Frame semantics and the nature of language},
  author={Fillmore, Charles J},
  journal={Annals of the New York Academy of Sciences},
  volume={280},
  number={1},
  pages={20--32},
  year={1976},
  publisher={Wiley Online Library}
}

@inproceedings{gimpel2010softmax,
  title={Softmax-margin CRFs: Training log-linear models with cost functions},
  author={Gimpel, Kevin and Smith, Noah A},
  booktitle={Human Language Technologies: The 2010 Annual Conference of the North American Chapter of the Association for Computational Linguistics},
  pages={733--736},
  year={2010}
}

\appendix



\end{document}